%% file: main.tex
\documentclass[10pt,twocolumn,letterpaper]{article}

\usepackage{cvpr}              


\definecolor{cvprblue}{rgb}{0.21,0.49,0.74}
\usepackage[pagebackref,breaklinks,colorlinks,allcolors=cvprblue]{hyperref}
\usepackage[linesnumbered,ruled,vlined]{algorithm2e}
\SetKwInput{KwInput}{Input}
\SetKwInput{KwOutput}{Output}

\title{How to Achieve Higher Accuracy with Less Training Points?}

\author{
Jinghan Yang, \ Anupam Pani, \ Yunchao Zhang\\
HKU Musketeers Foundation Institute of Data Science\\
The University of Hong Kong\\
{\tt\small \{eciel,apani3,yunchaoz\}@connect.hku.hk}
}

\begin{document}
\maketitle



\begin{abstract}
In the era of large-scale model training, the extensive use of available datasets has resulted in significant computational inefficiencies. To tackle this issue, we explore methods for identifying informative subsets of training data that can achieve comparable or even superior model performance. We propose a technique based on influence functions to determine which training samples should be included in the training set. We conducted empirical evaluations of our method on binary classification tasks utilizing logistic regression models. Our approach demonstrates performance comparable to that of training on the entire dataset while using only 10\% of the data. Furthermore, we found that our method achieved even higher accuracy when trained with just 60\% of the data. \footnote{This work is for the course project of DATA8009 at HKU IDS and has not been reviewed.
}
\end{abstract}

\section{Introduction}
Current model training paradigms typically rely on exhaustive utilization of entire datasets, resulting in significant resource inefficiencies, including excessive computational power consumption, energy expenditure, and carbon emissions \cite{ahmad2022data}. Such practices are at odds with the principles of Green AI, which emphasize environmentally sustainable and resource-efficient machine learning \cite{schwartz2020green, verdecchia2023systematic, barbierato2024toward}. Thus we consider the question: Can we achieve comparable or superior model performance while using fewer training samples?

To address this question, we turn to data-instance analysis methods, which provide insights into how individual training samples influence a model's predictions. 
These methods enable understanding of model behavior from a statical perspective \cite{pezeshkpour2021empirical, gu2023iaeval}. Such techniques have found widespread application in diverse domains, including finance, economics, and public policy, where they support decision-making and improve transparency in machine learning systems \cite{bracke2019machine, broderick2020automatic, amarasinghe2023explainable}. 
By analyzing how models make predictions on training points, researchers can debug predictions, denoise data, assess the robustness of models and training sets, and uncover biases within the training data \cite{IF, adebayo2020debugging, han2020explaining, teso2021interactive}.
The influence function is an effective tool for identifying training points significantly impacting parameter changes or loss for a specific test point \cite{cook,IF}. 
Recent work by \cite{yang-etal-2023-many, yang-etal-2024-relabeling} has extended this approach, proposing methods to flip test predictions by  removing or relabeling training samples. While these efforts focus on analyzing the robustness of models and datasets, the implications of such modifications on overall model performance remain unexplored.

Building on the influence functions and their application in modifying test predictions, we propose a novel metric to quantify the influence of adding training samples on test predictions. This metric enables us to determine whether a training sample has a positive or negative influence on the model's ability to make correct predictions. To evaluate our approach, we focus on a simple setting: a logistic regression model for binary classification tasks. Within this framework, we propose and compare several heuristic methods for selecting training subsets that achieve comparable or superior performance with significantly fewer training samples.

In conclusion, our work makes the following contributions:
\begin{enumerate}
\item Method to determine the influence of adding a training point to a test prediction by influence function.
\item Heuristic methods for training subset selection based on model behavior and data attributes.
\item Empirical validation demonstrates that our method achieves comparable accuracy using only 10\% of the training samples and performs even better with just 60\% of the dataset.
\end{enumerate}

\section{Background}

\noindent{\bf Influence Function:}
Influence functions are classical tools from robust statistics that measure the effect of infinitesimal perturbations to the training data on the learned model parameters and predictions \cite{hampel1974influence, cook1980characterizations, cook}. In machine learning, they have been employed to identify training instances that significantly impact a specific test prediction \cite{IF}. By analyzing these influential training points, researchers can debug models by uncovering mislabeled data, outliers, and biases in the training set \cite{adebayo2020debugging, han2020explaining, teso2021interactive}.

Extensions of influence functions have been proposed to assess the impact of changes not only in the presence of training points but also in their features and labels. For example, \cite{warnecke2021machine} applied influence functions to measure the effect of altering training point features and labels, aiding in machine unlearning tasks. Similarly, \cite{kong2021resolving} utilized influence functions to identify and correct noisy training samples, enhancing model performance during training.

Building on these approaches, our research leverages influence functions to determine which subsets of training data should be added to improve the test accuracy. 

\vspace{0.5em}
\noindent{\bf Flipping Model Predictions:}
Several studies have explored methods for changing model predictions by manipulating training data. \cite{pmlr-v162-ilyas22a} analyzed how variations in training data affect overall model behavior, providing insights into model robustness. In the field of economic modeling, \cite{harzli2022minimal} investigated altering specific predictions by identifying the minimal informative feature sets necessary for those predictions. Research on counterfactual examples aims to explain model predictions by identifying critical feature values responsible for certain outcomes \cite{kaushik2019learning}. By measuring the influence of training points on test predictions, methods have been proposed to flip a test prediction by removing or modifying influential training data \cite{broderick2020automatic, yang-etal-2023-many}. Specifically, \cite{yang-etal-2024-relabeling} introduced techniques to flip a test prediction by relabeling certain training points.
However, these methods primarily focus on altering the prediction of a single test point. 
Our research aims to investigate the impact of modifying training subsets to alter a group of test predictions. By doing so, we can improve overall model performance through targeted training data modifications.


\section{Preliminaries}
\label{sec:preliminaries}

In this section, we introduce the notation and foundational concepts that underpin our work, focusing on how influence functions can be used to understand and alter model predictions by modifying training data.

\subsection{Problem Setup and Notation}

Consider a binary classification problem with a training dataset $\mathcal{Z}^{\text{tr}} = \{z_1, \ldots, z_N\}$, where each data point $z_i = (x_i, y_i)$ consists of features $x_i \in \mathcal{X}$ and a label $y_i \in \mathcal{Y} = \{0, 1\}$. We aim to learn a model $f_w: \mathcal{X} \rightarrow \mathcal{Y}$ parameterized by $w \in \mathbb{R}^p$. The model parameters are estimated by minimizing the empirical risk:
\begin{equation}
\hat{w} = \underset{w}{\operatorname{argmin}}\, \mathcal{R}(w) = \underset{w}{\operatorname{argmin}} \left( \frac{1}{N} \sum_{i=1}^N \mathcal{L}(z_i, w) + \frac{\lambda}{2} \|w\|^2 \right),
\label{eq:empirical-risk}
\end{equation}
where $\mathcal{L}(z_i, w)$ is the loss function measuring the discrepancy between the predicted and true labels for $z_i$, and $\lambda$ is a regularization hyperparameter.

We assume that $\mathcal{R}(w)$ is twice differentiable and strongly convex with respect to $w$. The Hessian matrix of the empirical risk at $\hat{w}$ is given by:
\begin{equation}
H_{\hat{w}} = \nabla_w^2 \mathcal{R}(\hat{w}) = \frac{1}{N} \sum_{i=1}^N \nabla_w^2 \mathcal{L}(z_i, \hat{w}) + \lambda I,
\end{equation}
where $I$ is the identity matrix.

Suppose we add a new subset from an additional training set $\mathcal{S} \subset \mathcal{Z}^{\text{tr}'}$ to continue to train the model. The new parameters $\hat{w}_{\mathcal{S}}$ after training is defined by:
\begin{equation}
\hat{w}_{\mathcal{S}} = \underset{w}{\operatorname{argmin}} \left( \mathcal{R}(w) + \frac{1}{N} \sum_{z_i \in \mathcal{S}} \mathcal{L}(z_i, w)) \right),
\label{eq:modified-risk}
\end{equation}

Due to the computational impracticality of exhaustively searching all possible subsets $\mathcal{S}$ by retraining the model, we seek efficient methods to estimate the influence of modifying training points on the test predictions. By leveraging influence functions, we can approximate the effect of changes in the training data without retraining the model for each possible subset.
Influence function is a classical tools from robust statistics used to measure the effect of perturbations on the training data on the estimated parameters and predictions \cite{IF}. Thus we use influence function to measure the change of a test probability without retraining the model.

The change in the model parameters $\Delta w = \hat{w}_{\mathcal{S}} - \hat{w}$ can be approximated using the influence function:
\begin{equation}
\Delta w \approx - H_{\hat{w}}^{-1} \left( \frac{1}{N} \sum_{z_i \in \mathcal{S}} \nabla_w \mathcal{L}(z_i, w)) \right).
\label{eq:delta-w}
\end{equation}

The influence of modifying $\mathcal{S}$ on the prediction for a test point $z_t = (x_t, y_t)$ can be estimated by the change in the predicted value:
\begin{equation}
\Delta f_t = f_{\hat{w}_{\mathcal{S}}}(x_t) - f_{\hat{w}}(x_t) \approx \nabla_w f_{\hat{w}}(x_t)^\top \Delta w.
\label{eq:delta-f}
\end{equation}

This expression, referred to as \textbf{IP-adding}, allows us to estimate how adding training points in $\mathcal{S}$ influences the prediction at $x_t$ without retraining the model.

\begin{algorithm}[!ht]
\caption{An Algorithm to Find Training Points to Add}
\label{alg:alg1}
\DontPrintSemicolon
\KwInput{
  $f$: Model; 
  $\mathcal{Z}^{\text{tr}}$: Original training set; 
  $N$: Number of total training points; 
  $\mathcal{Z}^{\text{tr}'}$: Additional training set; 
  $\mathcal{Z}^{\text{val}}$: Validation set; 
  $\hat{w}$: Parameters estimated from $\mathcal{Z}^{\text{tr}}$; 
  $\mathcal{L}$: Loss function; 
  $x_t$: A test point; 
  $\tau$: Classification threshold (e.g., 0.5)
}
\KwOutput{$\mathcal{S}$: Training subset to add } 
$H \leftarrow \nabla_{w}^2  \mathcal{L}(\mathcal{Z}^{\text{tr}}, \hat{w})$ \\
$\nabla_w l \leftarrow \nabla_{w} \mathcal{L}(\mathcal{Z}^{\text{tr}'} , \hat{w})$ \\
$\Delta w \leftarrow \frac{1}{N} H^{-1} \nabla_w l$ \\
$\Delta f \leftarrow \nabla_{w} f(\mathcal{Z}^{\text{val}})^\intercal \Delta w$ \\
$\mathcal{S} \leftarrow \{\}$ \\
\For{each training point $z_i \in \mathcal{Z}^{\text{tr}'}$}{
    $score \leftarrow 0$ \\
    \For{each validation point $z_j \in \mathcal{Z}^{\text{val}}$}{
        \If{$\text{sign}(\Delta f[i,j]) = \text{sign}(\tau - f[j])$}{
            $score \leftarrow score + \Delta f[i,j]$
        }
        \Else{
            $score \leftarrow score - \Delta f[i,j]$ 
        }
    }
    \If{$score > 0$}{
        $\mathcal{S} \leftarrow \mathcal{S} \cup \{z_i\}$ 
    }
}
\KwRet $\mathcal{S}$
\end{algorithm}


\section{Methodology}
Based on IP-adding,  we propose several methods to add additional training points based on their influence on the validation set points. We assume that the distribution of the test set and validation set is similar, then select training points based on validation set. We firstly detailed the  idea in Method 1 and described its variation in followed methods.

\textbf{Method 1}:
The algorithm \ref{alg:alg1} iterates over each point in the additional training set, $x_i \in \mathcal{Z}^{\text{tr}'}$, and evaluates its potential contribution to the model. For each point, the contribution is calculated by comparing the sign of the change in the model's output, $\Delta_t f$, with the model's predicted output for each validation point, $x_j \in \mathcal{Z}^{\text{val}}$. If the sign aligns (indicating that the inclusion of $x_i$ would help the model make a correct classification for $x_j$), the contribution is incremented; otherwise, it is decremented. If the overall contribution for a point $x_i$ is positive, it is added to the minimal training subset $\mathcal{S}_t$, which is returned as the final output. The algorithm aims to select the most influential examples from the additional training set to improve the model's performance on the test point $x_t$.



\subsection{Method 2: Weighted contribution based training point selection}
Building upon the foundation of Method 1, we introduce a weighting mechanism to refine the selection of additional training points. This method assigns a weight to each training point based on the discrepancy between the ground truth label and the model's predicted value for that point. This discrepancy $d_i$, for a sample $i$, is defined as:
\begin{equation}
    d_i = y_i - \hat{y_i}
\end{equation}
where $\hat{y_i}$ denotes the predicted value.

\subsection{Method 3: Threshold-based Selection with Top-K Prioritization}
In Method 3, we select training points by first applying a threshold to filter out those with negligible or negative cumulative influence on the validation set. The cumulative influence is calculated as the sum of prediction changes that move the model toward the correct labels. From the remaining points, we prioritize and select the top 10 percent with the highest cumulative influence. This approach ensures that only the most impactful training samples are added to improve test accuracy efficiently.

\subsection{Method 4: Flip based selection using influence scores} 
In Method 4, we select training points by evaluating their potential to flip predictions on the validation set. For each training point in the additional training set, we calculate its influence score and assess how many validation points it can flip to the correct label. Training points that cause the greatest number of prediction flips are prioritized for inclusion.

\subsection{Method 5: Penalizing Misaligned Samples}  
In Method 5, we adjust the selection of training points by assigning weights based on the alignment between the influence score and the direction of prediction correction. Specifically:
\begin{itemize}
    \item Misaligned samples (where the sign of the prediction change does not match the desired correction) are penalized more heavily with larger weights.
    \item Aligned samples (where the sign matches) are assigned smaller weights. 
\end{itemize}

\subsection{Method 6: Threshold-based contribution calculation} Previous contribution calculation only uses the influence scores without considering the threshold. However, it may not be the best selection. For example, let's suppose there are two data points $i$ and $j$ in the validation set, where the predicted values and ground truth values are $\hat{y_i} = 0.9, y_i=1.0$, $\hat{y_j} = 0.4, y_j=1.0$, and the threshold is $0.5$. If an additional training point can increase $\hat{y_j}$ from $0.4$ to $0.6$ while decreasing $\hat{y_i}$ from $0.9$ to $0.6$, it can increase the over all accuracy even if its contribution is negative ($-0.1$ using Method 1). In light of this, we take the threshold into account: we ignore the part above the threshold when adding the influence score and make flips, and ignore the influence that does not cause flips.

\section{Experiments}
This section demonstrates the model, dataset and settings for baselines.

\noindent{\bf Models}: We use a logistic regression model to satisfy the assumption of convex loss in the influence function.

\vspace{0.5em}
\noindent{\bf Datasets}:
We perform sentiment analysis using the Movie Review Sentiment dataset, specifically the binarized version of the Stanford Sentiment Treebank
\cite{socher2013recursive}. To fit the logistic regression model, we transform the text data into a Bag-of-Words representation.

\subsection{Baseline Settings}
We train the original model on original training set $\mathcal{Z}^{\text{tr}}$ (4152 training points).
Then use methods to select training samples from additional subset $\mathcal{Z}^{\text{tr}’}$ (2768 training points) based on their influence on the validation set, $\mathcal{Z}_{\text{val}}$. The goal is to select $k$ examples that will, when added to the training set, improve model performance the most. We compare our model train on selected data with following settings on a test set (different from the validation set):

\noindent{\textbf{Original:}} We use only the original training set samples, $\mathcal{Z}_{tr}$, for model training, without incorporating any additional samples.

\noindent{\textbf{Add Full:}} We use both the original training set, $\mathcal{Z}_{tr}$, and all the additional samples, $\mathcal{Z}^{\text{tr}’}$, for model training.

\noindent{\textbf{Random Selection}}
As a comparison, we also implement a random selection process where $k$ samples are randomly drawn from $\mathcal{Z}^{\text{tr}’}$ without considering their influence on the validation set.

\subsection{Results}

\begin{table}[h!]
\small 
\setlength{\tabcolsep}{3.5pt} 
\renewcommand{\arraystretch}{1.1} 
\caption{Performance comparison of influence-based sample selection methods versus baselines on Stanford Sentiment Treebank using logistic regression.}
\label{tab:methods_comparison}
\centering
\begin{tabular}{lcccc}
\toprule
\textbf{Variants} & \textbf{Added Percentage} & \textbf{Random} & \textbf{Val Acc} & \textbf{Test Acc} \\ \midrule
Original   & 0\%     & *     & 0.616   & 0.614   \\
Add Full   & 100\%  & *     & 0.623   & 0.627   \\
Method 1   & \textbf{60\%}  & 0.620 & 0.649   & \textbf{0.644} \\
Method 2   & 56\%  & 0.620 & 0.65 & 0.634   \\
Method 3   & \textbf{10\%}   & 0.617 & 0.650   & \textbf{0.630}   \\
Method 4   & 1\%    & 0.617 & 0.632   & 0.613   \\
Method 5   & 0.1\%     & 0.616 & 0.618   & 0.613   \\
Method 6   & 60\%  & *     & 0.639   & 0.643   \\ 
\bottomrule
\end{tabular}
\end{table}

Tab.~\ref{tab:methods_comparison} presents the performance of various variants of our method, compared to the original model and the "Add Full" setting (i.e., all additional training data are used). Notably, our proposed method achieves superior performance with significantly fewer additional training points. For instance, Method 3 utilizes only about $10\%$ of the additional training data yet achieves higher accuracy than Add Full on both the validation and test sets. This result highlights the effectiveness of selectively adding training data, demonstrating substantial improvements in both accuracy and efficiency. Furthermore, the results of Random show that randomly selecting additional training data performs worse than using all available data. On average, the random selection strategy results in a $3\%$ performance drop compared to our proposed selection mechanisms.

Among all the variants, Method 1 turns out to have the highest test accuracy. Method 1 has the most general and simplest mechanism, fairly calculating the contribution of each data point in the additional training samples. It selects points that have positive contributions as many as possible. 

\section{Conclusion}
In conclusion, our proposed method leverages a modified influence function to efficiently select training points, offering a practical solution to the challenge of training large models with exhausted data. Using only a fraction of the available training set, 10\% in binary classification, our approach achieves performance comparable to full dataset training, while significantly reducing computational cost. Furthermore, when training with 60\% of the data, our method improves accuracy, demonstrating its potential to improve the model performance with fewer data points. This work provides a promising direction for improving data quality in machine learning, especially in the era of large-scale models, and opens the door for future research on refining data selection techniques to optimize model training.

\section{Future Work}
This study suggests several future research : (1) Extending the methodology to complex neural architectures and multi-class scenarios, (2) Investigating the relationship between influence-selected training subsets and test distribution characteristics, (3) Developing theoretical frameworks to understand how sample selection impacts model generalization, (4) Exploring practical applications in resource-constrained learning scenarios, and (5) Investigating robustness to distribution shifts and label noise. These extensions would help advance both theoretical understanding and practical deployment of influence-based sample selection methods while maintaining the approach's computational advantages.

\bibliographystyle{IEEEtran}
\bibliography{main}

\end{document}


%% file: main.bbl
\begin{thebibliography}{10}
\providecommand{\url}[1]{#1}
\csname url@samestyle\endcsname
\providecommand{\newblock}{\relax}
\providecommand{\bibinfo}[2]{#2}
\providecommand{\BIBentrySTDinterwordspacing}{\spaceskip=0pt\relax}
\providecommand{\BIBentryALTinterwordstretchfactor}{4}
\providecommand{\BIBentryALTinterwordspacing}{\spaceskip=\fontdimen2\font plus
\BIBentryALTinterwordstretchfactor\fontdimen3\font minus \fontdimen4\font\relax}
\providecommand{\BIBforeignlanguage}[2]{{%
\expandafter\ifx\csname l@#1\endcsname\relax
\typeout{** WARNING: IEEEtran.bst: No hyphenation pattern has been}%
\typeout{** loaded for the language `#1'. Using the pattern for}%
\typeout{** the default language instead.}%
\else
\language=\csname l@#1\endcsname
\fi
#2}}
\providecommand{\BIBdecl}{\relax}
\BIBdecl

\bibitem{ahmad2022data}
T.~Ahmad, R.~Madonski, D.~Zhang, C.~Huang, and A.~Mujeeb, ``Data-driven probabilistic machine learning in sustainable smart energy/smart energy systems: Key developments, challenges, and future research opportunities in the context of smart grid paradigm,'' \emph{Renewable and Sustainable Energy Reviews}, vol. 160, p. 112128, 2022.

\bibitem{schwartz2020green}
R.~Schwartz, J.~Dodge, N.~A. Smith, and O.~Etzioni, ``Green ai,'' \emph{Communications of the ACM}, vol.~63, no.~12, pp. 54--63, 2020.

\bibitem{verdecchia2023systematic}
R.~Verdecchia, J.~Sallou, and L.~Cruz, ``A systematic review of green ai,'' \emph{Wiley Interdisciplinary Reviews: Data Mining and Knowledge Discovery}, vol.~13, no.~4, p. e1507, 2023.

\bibitem{barbierato2024toward}
E.~Barbierato and A.~Gatti, ``Toward green ai: A methodological survey of the scientific literature,'' \emph{IEEE Access}, vol.~12, pp. 23\,989--24\,013, 2024.

\bibitem{pezeshkpour2021empirical}
P.~Pezeshkpour, S.~Jain, B.~C. Wallace, and S.~Singh, ``An empirical comparison of instance attribution methods for nlp,'' \emph{arXiv preprint arXiv:2104.04128}, 2021.

\bibitem{gu2023iaeval}
P.~Gu, Y.~Shen, L.~Wang, Q.~Wang, H.~Wu, and Z.~Mao, ``Iaeval: A comprehensive evaluation of instance attribution on natural language understanding,'' in \emph{Findings of the Association for Computational Linguistics: EMNLP 2023}, 2023, pp. 11\,966--11\,977.

\bibitem{bracke2019machine}
P.~Bracke, A.~Datta, C.~Jung, and S.~Sen, ``Machine learning explainability in finance: an application to default risk analysis,'' 2019.

\bibitem{broderick2020automatic}
T.~Broderick, R.~Giordano, and R.~Meager, ``An automatic finite-sample robustness metric: When can dropping a little data make a big difference?'' \emph{arXiv preprint arXiv:2011.14999}, 2020.

\bibitem{amarasinghe2023explainable}
K.~Amarasinghe, K.~T. Rodolfa, H.~Lamba, and R.~Ghani, ``Explainable machine learning for public policy: Use cases, gaps, and research directions,'' \emph{Data \& Policy}, vol.~5, p.~e5, 2023.

\bibitem{IF}
P.~W. Koh and P.~Liang, ``Understanding black-box predictions via influence functions,'' in \emph{International conference on machine learning}.\hskip 1em plus 0.5em minus 0.4em\relax PMLR, 2017, pp. 1885--1894.

\bibitem{adebayo2020debugging}
J.~Adebayo, M.~Muelly, I.~Liccardi, and B.~Kim, ``Debugging tests for model explanations,'' \emph{arXiv preprint arXiv:2011.05429}, 2020.

\bibitem{han2020explaining}
X.~Han, B.~C. Wallace, and Y.~Tsvetkov, ``Explaining black box predictions and unveiling data artifacts through influence functions,'' \emph{arXiv preprint arXiv:2005.06676}, 2020.

\bibitem{teso2021interactive}
S.~Teso, A.~Bontempelli, F.~Giunchiglia, and A.~Passerini, ``Interactive label cleaning with example-based explanations,'' \emph{Advances in Neural Information Processing Systems}, vol.~34, pp. 12\,966--12\,977, 2021.

\bibitem{cook}
R.~D. Cook and S.~Weisberg, \emph{Residuals and influence in regression}.\hskip 1em plus 0.5em minus 0.4em\relax New York: Chapman and Hall, 1982.

\bibitem{yang-etal-2023-many}
\BIBentryALTinterwordspacing
J.~Yang, S.~Jain, and B.~C. Wallace, ``How many and which training points would need to be removed to flip this prediction?'' in \emph{Proceedings of the 17th Conference of the European Chapter of the Association for Computational Linguistics}, A.~Vlachos and I.~Augenstein, Eds.\hskip 1em plus 0.5em minus 0.4em\relax Dubrovnik, Croatia: Association for Computational Linguistics, May 2023, pp. 2571--2584. [Online]. Available: \url{https://aclanthology.org/2023.eacl-main.188}
\BIBentrySTDinterwordspacing

\bibitem{yang-etal-2024-relabeling}
\BIBentryALTinterwordspacing
J.~Yang, L.~Xu, and L.~Yu, ``Relabeling minimal training subset to flip a prediction,'' in \emph{Findings of the Association for Computational Linguistics: EACL 2024}, Y.~Graham and M.~Purver, Eds.\hskip 1em plus 0.5em minus 0.4em\relax St. Julian{'}s, Malta: Association for Computational Linguistics, Mar. 2024, pp. 1085--1098. [Online]. Available: \url{https://aclanthology.org/2024.findings-eacl.73}
\BIBentrySTDinterwordspacing

\bibitem{hampel1974influence}
F.~R. Hampel, ``The influence curve and its role in robust estimation,'' \emph{Journal of the american statistical association}, vol.~69, no. 346, pp. 383--393, 1974.

\bibitem{cook1980characterizations}
R.~D. Cook and S.~Weisberg, ``Characterizations of an empirical influence function for detecting influential cases in regression,'' \emph{Technometrics}, vol.~22, no.~4, pp. 495--508, 1980.

\bibitem{warnecke2021machine}
A.~Warnecke, L.~Pirch, C.~Wressnegger, and K.~Rieck, ``Machine unlearning of features and labels,'' \emph{arXiv preprint arXiv:2108.11577}, 2021.

\bibitem{kong2021resolving}
S.~Kong, Y.~Shen, and L.~Huang, ``Resolving training biases via influence-based data relabeling,'' in \emph{International Conference on Learning Representations}, 2021.

\bibitem{pmlr-v162-ilyas22a}
\BIBentryALTinterwordspacing
A.~Ilyas, S.~M. Park, L.~Engstrom, G.~Leclerc, and A.~Madry, ``Datamodels: Understanding predictions with data and data with predictions,'' in \emph{Proceedings of the 39th International Conference on Machine Learning}, ser. Proceedings of Machine Learning Research, K.~Chaudhuri, S.~Jegelka, L.~Song, C.~Szepesvari, G.~Niu, and S.~Sabato, Eds., vol. 162.\hskip 1em plus 0.5em minus 0.4em\relax PMLR, 17--23 Jul 2022, pp. 9525--9587. [Online]. Available: \url{https://proceedings.mlr.press/v162/ilyas22a.html}
\BIBentrySTDinterwordspacing

\bibitem{harzli2022minimal}
O.~E. Harzli, B.~C. Grau, and I.~Horrocks, ``Minimal explanations for neural network predictions,'' \emph{arXiv preprint arXiv:2205.09901}, 2022.

\bibitem{kaushik2019learning}
D.~Kaushik, E.~Hovy, and Z.~C. Lipton, ``Learning the difference that makes a difference with counterfactually-augmented data,'' \emph{arXiv preprint arXiv:1909.12434}, 2019.

\bibitem{socher2013recursive}
R.~Socher, A.~Perelygin, J.~Wu, J.~Chuang, C.~D. Manning, A.~Y. Ng, and C.~Potts, ``Recursive deep models for semantic compositionality over a sentiment treebank,'' in \emph{Proceedings of the 2013 conference on empirical methods in natural language processing}, 2013, pp. 1631--1642.

\end{thebibliography}
